\documentclass{article}
\usepackage{ijcai22}

\usepackage{times}
\usepackage{multirow}
\usepackage{soul}
\usepackage{url}
\usepackage[hidelinks]{hyperref}
\usepackage[utf8]{inputenc}
\usepackage[small]{caption}
\usepackage{graphicx}
\usepackage{amsmath}
\usepackage{amsthm}
\usepackage{booktabs}
\usepackage{algorithm}
\usepackage{algorithmic}
\title{Sparse Message Passing Network with Feature Integration for Online Multiple Object Tracking}

\author{
Bisheng Wang$^{1,2}$
\and
Horst Possegger$^2$\and
Horst Bischof$^{2}$\And
Guo Cao$^1$
\affiliations
$^1$School of Computer Science and Engerring, Nanjing University of Science and Technology, China\\
$^2$Institute of Computer Graphics and Vision, Graz University of Technology, Austria
\emails
316106002478@njust.edu.cn,
\{possegger, bischof\}@icg.tugraz.at,
caoguo@njust.edu.cn
}

\begin{document}

\maketitle

\begin{abstract}
Existing \textit{Multiple Object Tracking} (MOT) methods design complex architectures for better tracking performance. However, without a proper organization of input information, they still fail to perform tracking robustly and suffer from frequent identity switches. In this paper, we propose two novel methods together with a simple online \textit{Message Passing Network} (MPN) to address these limitations. First, we explore different integration methods for the graph node and edge embeddings and put forward a new IoU (Intersection over Union) guided function, which improves long term tracking and handles identity switches. Second, we introduce a hierarchical sampling strategy to construct sparser graphs which allows to focus the training on more difficult samples. Experimental results demonstrate that a simple online MPN with these two contributions can perform better than many state-of-the-art methods. In addition, our association method generalizes well and can also improve the results of private detection based methods.
\end{abstract}

\section{Introduction}
\label{sec:intro}

Multiple Object Tracking (MOT) is a complex but significant task in computer vision, which can be applied to different applications ranging from autonomous vehicles to automated video analysis and surveillance. Most MOT methods follow the \textit{tracking by detection} paradigm and split the task into two steps: (1) Detection. The targets in each frame are detected by different detectors. 
(2) Association. The detections belonging to the same target across different frames are then associated to form a target's trajectory. This paper focuses on the second step,~\textit{i.e.} how to perform robust tracking with the given detections.

For the association, earlier methods only use appearance features or motion information, which have difficulties handling crowded scenes. Some recent methods~\cite{braso2020learning,kim2021discriminative} begin to construct graph based methods. They take detections in each frame as nodes and assign the connections between detections as edges. Then the constructed graph is fed into graph neural networks to learn a better association. However, these graph based methods mainly focus on the design of networks, without considering the limitations of the constructed graphs. For example, these methods only coarsely feed the features extracted from detections into the networks, ignoring the importance of embedding temporal cues into graph. In addition, existing association methods commonly train all the connected pairs (\emph{i.e.} edges in the graph) simultaneously. Most of these pairs, however, are easy samples. During training, the convergence may drift to these easy and redundant samples.

To address these shortcomings of previous graph-based tracking approaches, we propose two novel improvements.
First, we integrate temporal cues for the embeddings of graph nodes and edges, which is beneficial for long term tracking, especially for occluded targets. We compare three different operations to identify the best integration method for robust tracking, including a new IoU (intersection over union) guided function. This new function does not need any training and has a superior performance, even surpassing LSTM based integration modules.

Second, experimental analysis shows that appearance features or IoU alone can correctly match most trajectory and detection pairs. To this end, instead of simply picking a fixed number of the nearest neighbours for each node, which is a common way for many graph based methods, we propose a hierarchical method to construct graph edges. The new graph construction uses a distance ratio test~\cite{lowe2004distinctive} to connect easy nodes first. Then other uncertain nodes are connected in a redundant way. This results in a much sparser graph, which reduces computation and helps to focus the training on the more difficult samples.
By adopting a simple online message passing network (MPN), the two operations realize a robust and accurate tracking system. 


To summarize, our contributions are as follows:

(1) We propose an IoU guided feature integration function, which incorporates temporal cues for robust tracking.

(2) A hierarchical method is introduced to construct a sparse but well-performing graph for MOT.

(3) Our method obtains a better performance than state-of-the-art online MOT methods for both public and private input detections.

\section{Related Work}
\label{sec:related}

We focus this summary on well-performing association-based trackers. To this end, we introduce related work based on the cues and architectures they use.

\textbf{Tracking with different cues.} Earlier methods use simple architectures and explore the performance of different cues. SORT~\cite{bewley2016simple} relied on a better detector~\cite{ren2015faster} and only took advantage of coordinate information. DeepSORT~\cite{wojke2017simple} combined location with appearance information and proposed a cascaded association strategy. 
Recent approaches~\cite{braso2020learning,papakis2020gcnnmatch,sadeghian2017tracking} combined different cues in one network for more robust performance.

\textbf{Tracking based on long term dependencies.} Using temporal information is a practical way for robust MOT. DeepSORT~\cite{wojke2017simple} saved the appearance features of all frames for each identity and chose the most similar trajectory for new detections. RNN\_LSTM~\cite{milan2017online} first investigated recurrent neural networks for MOT. They used the motion information to construct a tracking model, which exhibited an unsatisfactory performance. Notably better performance can be achieved by leveraging a structure of RNNs~\cite{sadeghian2017tracking}, which integrate appearance, motion and interaction information.
There are also some methods~\cite{saleh2021probabilistic,kim2021discriminative} which designed even more complex architectures to integrate temporal cues but only obtain minor improvements. We also leverage temporal information in this work. However, with a new integration function and a simple graph neural network, we obtain better tracking performance. 

\textbf{Tracking as a graph problem.} Many recent methods formulate the association step as a graph problem, where each detection is a node while edges indicate possible links among targets. Because of the simplicity, the Hungarian algorithm~\cite{kuhn1955hungarian} is most widely used for association. Recently, several works started to design complex trainable graph neural networks. MPNTrack~\cite{braso2020learning} adopted MPN~\cite{gilmer2017neural} with a time aware mechanism to train an offline tracking network. GCNNMatch~\cite{papakis2020gcnnmatch} introduced the graph convolutional network as their association network. GSM~\cite{liu2020gsm} proposed a novel graph representation together with a  matching strategy for efficient tracking. These works all focused on the graph neural network design. However, we put our attention on the temporal information integration and the construction of sparse graph.

\section{Method}
\label{sec:method}

First, we introduce our sparse message passing network in Section~\ref{sec:mpn}. Then, we explain how to integrate temporal cues and construct the sparse graph in Sections~\ref{sec:FeatureIntegration} and~\ref{sec:graph}, respectively. Finally, we explain our inference stage in Section~\ref{sec:inference}.

\subsection{Online Message Passing Network for MOT}
\label{sec:mpn}

\begin{figure}[t]
\begin{center}
\includegraphics[width=0.9\linewidth]{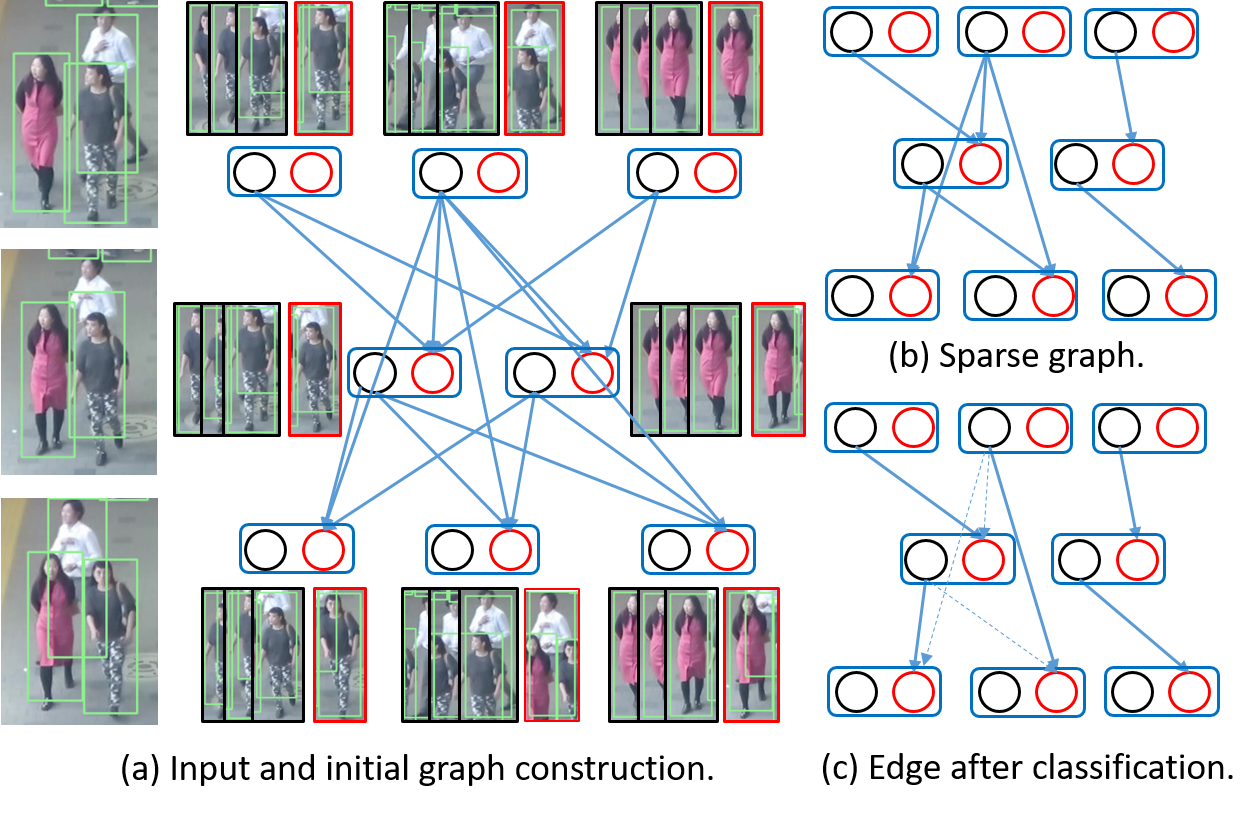}
\end{center}
\caption{Overview of our method. (a) The detected boxes and the initial graph. Green boxes in the image patches are the detections. Black and red circles represent the embeddings of trajectories and detections. Edges are directed and are always connected from trajectory embeddings to detection embeddings. (b) Our constructed sparse graph. (c) Association result after the MPN. Dashed lines are unmatched connections after classification.}

\label{fig:1_graph}
\end{figure}

\textbf{Graph construction.} 
In the MOT task, we are given a set of detections $\mathcal{D} = \{D_1, D_2, ..., D_N\}$, where $N$ is the number of detections across all input frames. Each detection $D_j = \{f_j, p_j, t_j\}, j \in \{1, 2, ..., N\}$, is represented by its extracted feature $f_j$, bounding box position $p_j$ and timestamp $t_j$. $D_j$ corresponds to the red boxes and circles in Figure~\ref{fig:1_graph}(a). 
At frame $t$, the task is to associate the corresponding detections $\mathcal{D}^{t} = \{D_j \in \mathcal{D} | t_j = t\}$ to the previous trajectories $\mathcal{T}^{t-1}$. This forms the corresponding trajectory set $\mathcal{T}^{t} = \{T_1, ..., T_M\}$, where M is the number of active trajectories in frame $t$. $T_i, i \in \{1, 2, ..., M\}$, represents a trajectory which integrates the historical information of object $i$ and corresponds to the black boxes and circles of Figure~\ref{fig:1_graph}(a).


With the trajectory set $\mathcal{T}^{t}$ and detection set $D^{t+1}$, we model the association of detections to trajectories as a bipartite graph $G=(\mathcal{T}^{t}, \mathcal{D}^{t+1}, E)$, where $\mathcal{T}^{t}$ denotes the active trajectories at time $t$, $\mathcal{D}^{t+1}$ are the detections for the new frame at time $t+1$ and $E \subset \mathcal{T}^{t} \times \mathcal{D}^{t+1}$ denotes the edges.
Each edge $e_{i, j} = (T_i, D_j)$, with $T_i \in \mathcal{T}^{t}$ and $D_j \in \mathcal{D}^{t+1}$, is directed, \emph{i.e.} $e_{i, j}$ always associates from existing trajectories to the detections. Our final goal is to keep only those edges where $T_i$ and $D_j$ have the same object identity and simultaneously to satisfy that each trajectory is connected to at most one detection and each detection is connected to at most one trajectory.



\textbf{Feature initialization.} Similar to many methods which use appearance features for better association, we input the bounding box images into a convolutional neural network to extract the appearance features.
Then, for each edge $e_{i, j} = (T_i, D_j)$ in the graph, $D_j$ is initialized by the appearance features from the current frame directly, named $f_j$, while the embedding of $T_i$ is integrated from features in the current frame as well as history frames, denoted $F_i$.
Different integration methods will be discussed in Section~\ref{sec:FeatureIntegration}. For the initial embedding of $e_{i, j}$, we compute the box appearance, location and time distances following MPNTrack~\cite{braso2020learning}.

\textbf{Sparse graph.} Different from many other graph based methods which coarsely assign each node with its neighbouring nodes to form edges, we introduce a distance ratio test inspired by SIFT~\cite{lowe2004distinctive} in our work to filter out redundant edges and construct a sparse graph, as shown in Figure~\ref{fig:1_graph}(b).
The specific procedure and benefits of this sparse graph will be elaborated in Section~\ref{sec:graph}.

\textbf{Neural message passing.} After constructing the graph with its initialized node and edge features, we feed it into a MPN~\cite{gilmer2017neural} to update the features.

Let $T_i^l$, $D_j^l$ and $h_{(i,j)}^{l}$ be the node and edge embeddings in the $l^{\text{th}}$ layer of the MPN. We divide the propagation procedure into two steps: One from nodes ($v$) to edges ($e$), and the other one from edges to nodes. These two steps are performed in turn for $L$ iterations as follows:
\begin{align}
    (v\ \rightarrow\ e)\quad &h_{i, j}^{l} =  \Psi_e([T_i^{l-1}, D_j^{l-1}, h_{(i, j)}^{l-1}]), \\
    \label{equ:edge function}
    (e\ \rightarrow\ v)\quad &x_{i, j}^{l} =  \Psi_v([T_i^{l-1}, h_{(i, j)}^{l}]),\\
    &T_i^{l} = \Phi({x_{(i, j)}^{(l)}}_{j \in N_i}),\\
    &y_{i, j}^{l} = \Psi_v([D_j^{l-1}, h_{(i, j)}^{l}]),\\
    &D_j^{l} = \Phi({y_{(i, j)}^{(l)}}_{i \in N_j}),
\end{align}%
where $\Psi_e$ and $\Psi_v$ are learnable functions shared across the entire graph. $\Phi$ is an order-invariant function, \textit{e.g.} a summation or an average.
The procedure follows the basic framework from~\cite{gilmer2017neural}, except that we update the node embeddings $T_i^l$, $D_j^l$ separately to ensure a bipartite graph. 

\textbf{Training.} For each iteration of training, we take a sequence of 20 continuous frames as one batch. Every time we take detections from one frame and pick the closest trajectories in previous frames for them to form a graph. Then the graph is propagated through the MPN. The final edge embeddings output from MPN are used to perform binary classification. For training, the binary cross-entropy loss is adopted with the weight parameter $\omega$ to adjust the imbalance between active and inactive edges.


Note that MPNTrack~\cite{braso2020learning} also adopts MPN as their basic architecture. However, we differ fundamentally from MPNTrack: (1) MPNTrack is an offline method which needs information from future frames, whereas our method is an online tracking framework. (2) The main work in MPNTrack focuses on how to design the tracking network, while our method considers how to integrate features and construct a sparse but efficient graph.

\subsection{Temporal Cues for Long Term Association}
\label{sec:FeatureIntegration}

The reasons for integrating temporal information are obvious: (1) In occlusion scenarios, temporal information can help prevent identity switches for partially occluded targets. (2) Leveraging temporal information reduces the distraction of noisy detections and thus leads to a more robust performance.

\begin{figure}[t]
\begin{center}
\includegraphics[width=0.95\linewidth]{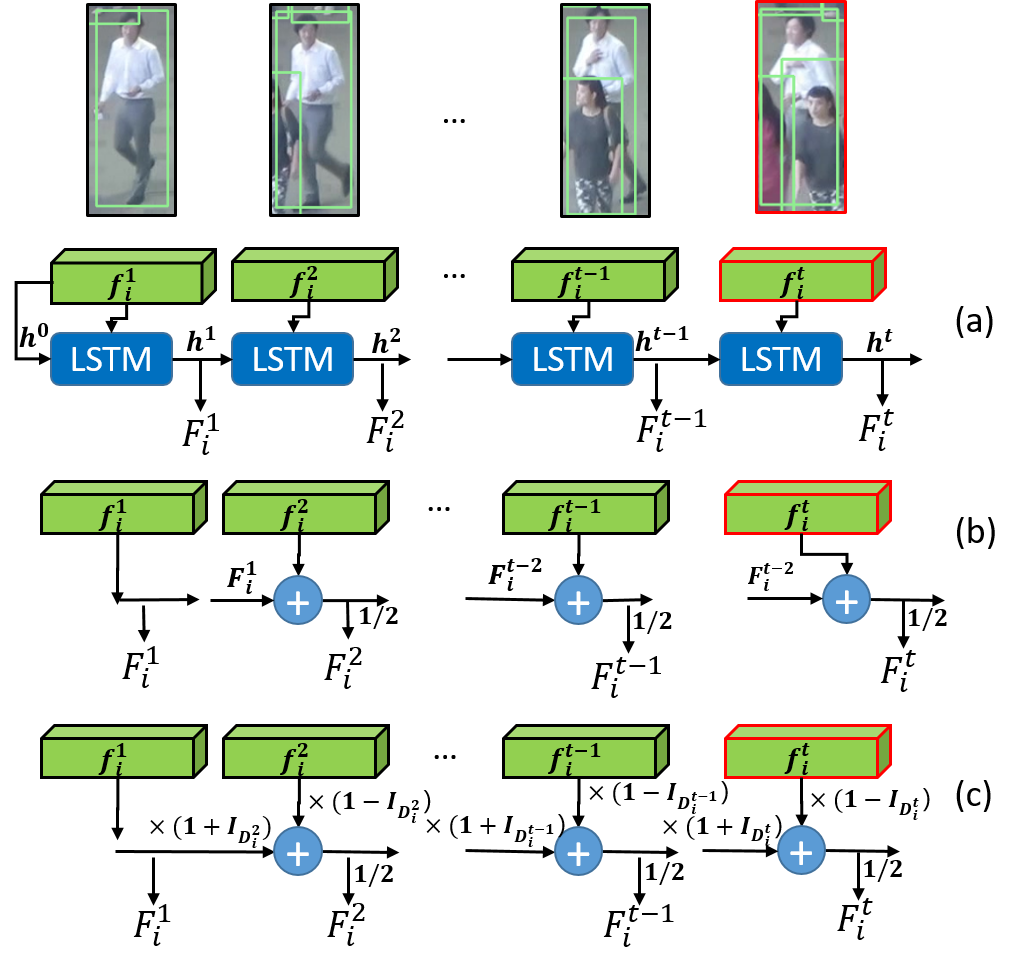}
\end{center}
\caption{Different methods to incorporate temporal context: (a) LSTM based module. (b) Average features. (c) Our IoU guided integration function. $I_{D_i^t}$ is the largest IoU that detection $D_i^t$ has with other targets.}

\label{fig:2_integration}
\end{figure}

To properly integrate the temporal information for trajectory $T_i$, we compare the following three different ways to compute the node embeddings, which is shown in Figure~\ref{fig:2_integration}. 

\textbf{LSTM based method.} For each trajectory node $T_i$ in frame $t$, we insert an LSTM module to integrate features before the MPN, as can be seen in Figure~\ref{fig:2_integration}(a). The feature $F_i^{t}$ for trajectory node $T_i^t$ is

\begin{equation}
    F_i^{t} = \operatorname{LSTM}(h_i^{t-1},\ f_i^{t}).
\end{equation}%

\textbf{Average features.} The second integration method, which is shown in Figure~\ref{fig:2_integration}(b), simply computes the average feature for a trajectory embedding. Given the trajectory embedding $F_i^{t-1}$ for $T_i$ in frame $t-1$ and the detection feature $f_i^t$ in frame $t$, its embedding $F_i^{t}$ is updated as
\begin{equation}
    F_i^{t} = 0.5 * (F_i^{t-1} + f_i^t).
\end{equation}%

\textbf{IoU guided features.} When computing the embedding of a trajectory $T_i$, if $T_i$ overlaps with other targets, we aim to take more information from its past frames. Otherwise,~\textit{i.e.}~if there is no overlap with other objects, we can increase our trust in the appearance information of the current frame. This strategy provides more history information in occlusion regions, which helps to reconnect targets after occlusion. Therefore, we propose an IoU guided integration function for the trajectory nodes, as is shown in Figure~\ref{fig:2_integration}(c). For each target $D_i^{t}$ in frame $t$, we compute its IoU with all other targets in the same frame. Then the IoU guided feature integration function is computed as
\begin{equation}
    F_i^{t} = 0.5 * (F_i^{t-1} * (1 + I_{D_i^{t}}) + f_i^t * (1 - I_{D_i^{t}})),
\end{equation}%
where $I_{D_i^{t}}$ is the maximum overlap that detection $D_i$ has with other targets. $f_i^t$ is the feature of a trajectory in the new frame.

\subsection{Sparse Graph}
\label{sec:graph}

Existing graph based methods coarsely assign each node a fixed number of neighbouring nodes based on location distances to form the edges, which will inevitably cause many redundant edges. During training, the model will tend to converge to these easy redundant edges but does not handle the condition of occlusion and long term connection, which should be the prime objective. 


To avoid redundant edges, we adopt a hierarchical idea to construct edges. Among all the edges connected to trajectory $T_i$, if $T_i$ is confident about which one of the edges is the true connection, then we will drop all the other edges and only keep this confident one. Otherwise, we will follow the common way to connect $T_i$ to the neighbouring detections. 

To this end, we leverage the distance ratio test initially proposed for SIFT keypoint matching~\cite{lowe2004distinctive} to realize this idea. For each trajectory $T_i$, let the detections connected to it be $D_j,\ j\in\{1, 2, ..., M\}$, where M is the number of connections. We then compute the distances $\delta_{i,j}$ between $T_i$ and all its connected detections. Afterwards, we sort the distances and compare the two closest, \emph{i.e.} $\delta_{i,j}^{\text{min}}$ and $\delta_{i,j}^{\text{2nd}}$. If
\begin{equation}
\label{equ:lowe}
    \delta_{i,j}^{\text{min}} < \alpha * \delta_{i,j}^{\text{2nd}},\ \alpha \in (0, 1),
\end{equation}%
we take the edge to the detection with the minimum distance as the only connection for $T_i$ and remove all other edges. Here, $\alpha$ is a pre-defined parameter.

There are two options to compute the distances: appearance feature distance and IoU. We make an analysis to evaluate how the two options perform as follows. We apply Equation \eqref{equ:lowe} to all the trajectory nodes throughout training sequences. For any trajectory which satisfies Equation \eqref{equ:lowe}, if the kept edge connects the same identity, we increment the number of true matches. If the kept edge connects different identities, we count it as false match. Trajectories which do not satisfy the inequality are considered inconclusive matches. The result of this analysis on the MOT17 training dataset is shown in Table~\ref{tab:lowe}.

From Table~\ref{tab:lowe}, we can conclude the following points: (1) The number of true matches (T) indicates that most easy connections can be classified directly using appearance features or IoU without graph training. (2) Using appearance feature distance causes significantly less false matches than using IoU distance. We will also compare how they perform during tracking in Section~\ref{sec:ablation}. (3) Applying the distance ratio test generates a much sparser graph, so that the training will be easier and optimization can focus on the more difficult edges, \emph{i.e.} association examples..

\begin{table}
\centering
\caption{The number of True (T), False (F) and Inconclusive (I) matches after applying the ratio test with IoU ($R_{\text{IoU}}$) and appearance feature distance ($R_{\text{app}}$).}
\label{tab:lowe}
\resizebox{0.95\columnwidth}{!}{
\begin{tabular}{llrrrrr}
\hline
                & $\alpha$     & 0.2    & 0.3   & 0.4   & 0.5   & 0.6    \\
\hline
$R_{\text{IoU}}$           & T        & 66350   & 77771  & 86526 & 92548 & 96864     \\
                & F       & 864     & 980    & 1096  & 1192  & 1274      \\
                & I        & 35918   & 24497  & 15742 & 9720  & 5404       \\
\hline
$R_{\text{app}}$            & T       & 20912   & 53904 & 77581 & 90173 & 96566    \\
                & F       & 0       & 1     & 5     & 31    & 72         \\
                & I        & 81886   & 48894 & 25217 & 12625 & 6232     \\
\hline
\end{tabular}
}
\end{table}

\begin{algorithm}[tb]
\caption{Pseudo-code of inference stage}
\label{alg:inference}
\textbf{Input}: Sequence $S$, detections $\mathcal{D}$, Kalman Filter (KF)\\
\textbf{Parameter}: Ratio parameter $\alpha$, Score threshold $\tau$\\
\textbf{Output}: Trajectories $\mathcal{T}$
\begin{algorithmic}[1] 
\STATE Initialization: $\mathcal{T} \leftarrow \emptyset$
\FOR{frame $t$ in Sequence S}
\STATE $\mathcal{D}^t \leftarrow \mathcal{D}$.
\IF {$\mathcal{D}^t$ is $\emptyset$ or $\mathcal{T}^{t-1}$ is $\emptyset$ }
\STATE Graph $G$ is None.
\ELSE
\STATE Set all edges $E$ with nodes $\mathcal{T}^{t-1}$ and $\mathcal{D}^t$.
\STATE Apply Equation \eqref{equ:lowe} to $E$ and get sparse graph $G$.
\ENDIF
\STATE Pass $G$ into MPN to get scores for all edges.
\STATE Get ranked positive edges $E_{\text{pos}}$ with $\tau$.
\STATE Get matched pairs $M$, unmatched trajectories $\mathcal{T}^{t-1}_{\text{unmatched}}$ and unmatched detections $\mathcal{D}_{\text{unmatched}}$.
\FOR{$(T_i, D_j)$ in $M$}
\STATE $T_i \leftarrow D_j$.
\ENDFOR
\FOR{$D_j$ in $\mathcal{D}_{\text{unmatched}}$}
\STATE Initialize a new trajectory $\mathcal{T}^{t} = \mathcal{T}^{t-1} \bigcup D_j$.
\ENDFOR
\FOR{$T_i$ in $\mathcal{T}^{t-1}_{\text{unmatched}}$}
\STATE Forecast location of $T_i$ with KF and constraints.
\ENDFOR
\ENDFOR
\STATE \textbf{return} $\mathcal{T}$
\end{algorithmic}
\end{algorithm}

\subsection{Inference}
\label{sec:inference}

This section explains the test stage of our tracking method. The pseudo-code is listed in Algorithm \ref{alg:inference}.
In each frame $t$, we obtain the detections $\mathcal{D}^t$ and the corresponding features. With the tracked trajectories $\mathcal{T}^{t-1}$, we can construct edges between $\mathcal{T}^{t-1}$ and $\mathcal{D}^t$. After that, we apply the distance ratio test to filter out easy edges and construct a sparse graph $G$ (lines 2 to 9).

We initialize the node and edge features for $G$ and propagate $G$ through the MPN to get the confidence scores for all connected edges. Edges with scores lower than $\tau$ will be removed and the remaining edges will be ranked from high to low. We get the final connected edges based on the order of the scores to ensure each trajectory is only connected to at most one detection and each detection is also only connected to at most one trajectory. In this way, we get the matched pairs $M$, unmatched trajectories $\mathcal{T}^{t-1}_{\text{unmatched}}$ and unmatched detections $\mathcal{D}_{\text{unmatched}}$ (lines 10 to 12).

For matched edges $(T_i, D_j)$ in $M$, we update $T_i$ with the feature of $D_j$ using the integration function in Section~\ref{sec:FeatureIntegration}. For any unmatched detection, we initialize a new trajectory (lines 13 to 18). 

As for each unmatched trajectory, if the identity has been lost for more than a fixed number of frames (set as 80), we will delete it. Otherwise, we forecast a target's location using a Kalman Filter (KF). We apply the following constraints on these forecast locations to avoid false detections: (1) If more than half of a target's predicted bounding box is outside the field-of-view, we presume it has left the scene and stop forecasting. 
(2) We adopt DRT-net~\cite{wang2021drt} to verify the predicted boxes. DRT-net helps filter out false detections and refine imprecise true detections. Once the predicted box is classified as a false detection by DRT-net, indicating that the forecasting is not reliable anymore, we stop forecasting. This constraint can eliminate most of the false forecasts. (3) We compute the distance between appearance features of the predicted boxes and the trajectories. If the distance is beyond a pre-defined threshold, we will stop forecasting. In fast moving videos, the forecast boxes may drift from one target to another nearby target. This constraint is useful for such situations. All the three constraints are solely applied to predicted boxes, which only causes a negligible computational overhead (lines 19 to 21).


\section{Experiments}
\label{sec:experiments}

\subsection{Settings}
\label{sec:settings}

\textbf{Datasets.} We evaluate our method on the publicly available and widely used MOT16~\cite{milan2016mot16}, MOT17 and MOT20~\cite{dendorfer2020mot20} datasets from the MOTChallenge\footnote{\url{https://motchallenge.net}} benchmark. MOT16 and MOT17 use the same 7 training videos and 7 test videos but they differ in  the provided public detections and quality of the annotations. These videos are challenging due to their frequent occlusions, varied camera perspectives, human poses as well as complex backgrounds. MOT20 is a new benchmark which consists of 8 challenging sequences. Compared with MOT16 \& MOT17, MOT20 has more crowded scenes and complex backgrounds. Furthermore, the test set in MOT20 include another two unknown scenes which are different from the scenes in the training sets in order to measure the generalization capabilities of detectors and trackers.

\textbf{Evaluation metrics.} Same as most MOT methods, we evaluate and compare our method with other methods using the widely accepted CLEAR MOT metrics~\cite{bernardin2008evaluating} (including MOTA, IDF1), the trajectory quality measures~\cite{li2009learning} (which contain FP, FN, IDs) and the recently proposed HOTA~\cite{luiten2021hota}. MOTA, FP and FN are more related to the detection performance, while IDF1 and IDS evaluate the identity preservation ability and measure the association performance. HOTA is a new metric which provides a more balanced evaluation between detection and association performance.

\textbf{Implementation details.} 
To extract appearance features, we adopt the person re-identification model from~\cite{luo2019bag} using the same datasets as MPNTrack~\cite{braso2020learning}.
To train the MPN model, we set the batch size as 8 graphs for each iteration. Each graph is constructed with detections from 15 continuous frames, which are sampled at 6 frames per second for static camera sequences and 9 frames per second for those with a moving camera. We randomly remove the nodes in the graph and shift the detection boxes to do data augmentation. We run 25 epochs with an initial learning rate 0.001, which is divided by 10 every 7 epochs. The parameters are trained using the Adam optimizer with $\beta_1$ and $\beta_2$ set to 0.9 and 0.999, respectively.


\subsection{Ablation Studies}
\label{sec:ablation}

In this section, we conduct detailed experiments to evaluate the contribution of each proposed component. We follow MPNTrack~\cite{braso2020learning} to split MOT17 training data into different sets and do 3-fold cross-validation to evaluate the performance of different modules in our work. Tracktor~\cite{bergmann2019tracking} is chosen as a baseline method because we use the detections from Tracktor. To understand our method, we conduct the following experiments. First, we realize a vanilla MPN and insert the three integration functions from Section~\ref{sec:FeatureIntegration} into the vanilla MPN to compare whether these functions can help improve the performance. Second, we compare the two options for the distance ratio test to construct the sparse graph. Finally, we also show the performance of applying Kalman Filter with and without constraints to help forecast lost targets. All these comparisons are summarized in Table~\ref{tab:ablation}.  

\textbf{Integration methods.} We design the following experiments to show how the integration functions perform: First, we realize a vanilla MPN, where we only use the feature of the last assigned detection for $T_i$ without any feature integration (the first line of MPN in Table~\ref{tab:ablation}). Then, we insert the feature integration functions in Section~\ref{sec:FeatureIntegration} to update the node feature of $T_i$ before passing through the network. In this part, we do not construct a sparse graph, but keep a fixed number of edges (this parameter is set 20 and analyzed in the supplementary material) for each node based on the pair distances. The results are compared in Table~\ref{tab:ablation}.

As can be seen, the vanilla MPN already obtains a notably better association performance than the baseline Tracktor. It shows the benefits of MPN which can combine different information during training. Based on the vanilla MPN, applying different integration functions consistently achieves further improvements, indicating that integrating temporal cues yields a more robust performance by reducing frequent switches and, thus, preserving the identities. 
 
Out of the three integration functions, applying LSTM module surprisingly performs worst. We analyze the model and find that the integrated feature tends to be close to the feature from the most recently assigned detection. Therefore, it performs slightly worse in the occlusion regions, where the identities need longer history information to prevent drifting. This also affects the identity preservation, causing a relatively low IDF1 score. 
Our new proposed IoU guided function (IF) performs slightly better than the average function (AF) because it balances long term re-connection and short term switches.

\begin{table}
\centering
\caption{The performance of applying different modules of our method. FI, R and F represent feature integration, distance ratio test and forecasting, respectively. W/o c and w/ c in the Forecasting column mean forecasting without and with constraints. $\uparrow$/$\downarrow$ denote that higher/lower scores are better.}
\label{tab:ablation}
\resizebox{1.0\columnwidth}{!}{
\begin{tabular}{ccccrrr}
\hline
Module  &FI   &R   &F        & MOTA$\uparrow$      & IDF1$\uparrow$      & IDS$\downarrow$     \\
\hline
\multicolumn{4}{c}{Tracktor~\cite{bergmann2019tracking}}       & 67.8      & 66.8      & 735   \\
\hline
MPN    & $\times$         & $\times$       & $\times$   & 68.2      & 70.5     & 310     \\
        &LSTM       & $\times$       & $\times$     & 68.2      & 71.9      & 286    \\
        &AF         & $\times$       &  $\times$     & 68.2      & 72.3      & 272    \\
         &IF  &  $\times$     &  $\times$     & 68.2     & 72.7       & 257    \\
             &IF     &$R_{\text{IoU}}$      &  $\times$    & 68.2      & 69.9     & 354    \\
                &IF     &$R_{\text{app}}$    & $\times$      & 68.3      & 73.3      & \textbf{222}    \\
                &IF         &$R_{\text{app}}$   & w/o c  & 63.0  &70.5    & 315    \\
             &IF         &$R_{\text{app}}$   & w/ c  & \textbf{71.1}  &\textbf{74.3}    & 235    \\

\hline
\end{tabular}
}
\end{table}

\begin{table}
\centering
\caption{The average edge numbers and computation time with and without applying the ratio test to construct the sparse graph.}
\label{tab:computation}
\begin{tabular}{lrr}
\hline
          & edges      & time (ms)     \\
\hline
w/o $R$  & 397     & 36.6   \\
w/ $R_{\text{IoU}}$     & 81    & 31.9    \\
w/ $R_{\text{app}}$     & 208    & 32.6    \\
\hline
\end{tabular}
\end{table}

\textbf{Distance ratio test.} We make the following comparisons to analyze the performance of applying the distance ratio test to construct the sparse graph.  First, as seen in Table~\ref{tab:computation}, we compare the average number of graph edges and computation time before and after applying the distance ratio test to the graph.  Second, we compare the tracking results comparison with and without distance ratio test in Table~\ref{tab:ablation}. 
We compare the performance of the IoU-based ratio test ($R_{\text{IoU}}$) and appearance distance-based ratio test ($R_{\text{app}}$).
The parameter $\alpha$ for $R_{\text{IoU}}$ and $R_{\text{app}}$ is set to 0.1 and 0.3, respectively.

As shown in Table~\ref{tab:computation}, by using $R_{\text{IoU}}$ and $R_{\text{app}}$, we can cut around 80\% and 48\% of the edges in contrast to the dense graph, respectively. The speed is only a little bit faster because: (1) Cutting edges only save the computation of Equation (1). The computation of Equation (2)-(4) does not change. (2) The process of cutting edges also costs some time. However, the overall association stage is quite fast and can be applied into any tracking methods without adding too much computation.

As for the results, using $R_{\text{IoU}}$ has a relatively low IDF1 and more ID switches. This is in line with the evaluation results from Table~\ref{tab:lowe} because using the IoU distance creates more false connections. $R_{\text{app}}$, on the other hand, exhibits more improvements. In our experiments, we found out that with most of the easy edges removed by the ratio test, the MPN training can focus on the difficult ones and thus have a better convergence.

\textbf{Forecasting.} We show the performance of forecasting in the last three lines of Table~\ref{tab:ablation} with the following three experiments: (1) The first one is our MPN with IoU guided feature integration and $R_{\text{app}}$. (2) We then apply the Kalman filter without constraints to forecast hte position of lost targets. (3) We finally add the constraints from Section~\ref{sec:inference}. Forecasting the locations of lost targets with constraints increases the MOTA significantly by recovering the missing detections, which also slightly increases IDF1. Without constraints the performance is abysmal because forecasting causes many false detections (the second to last line of Table~\ref{tab:ablation}).


\begin{table}
\centering
\caption{Results of different public detection based methods. G and $\star$ denote graph based methods and offline methods, respectively. $\uparrow$/$\downarrow$ denote that higher/lower scores are better.}
\label{tab:public_det_results}
\resizebox{1.0\columnwidth}{!}{
\begin{tabular}{llrrrrrr}
\hline
Dataset     &Method              & MOTA$\uparrow$      & IDF1$\uparrow$      & HOTA$\uparrow$      &IDS$\downarrow$       &FP$\downarrow$     &FN$\downarrow$     \\
\hline
MOT16       &BLSM-MTP~\cite{kim2021discriminative}            & 48.3      & 53.5      & -         &733       &9799  &83712 \\
            &DeepMOT~\cite{xu2020train}            & 54.8      & 53.4      & 42.2         &645       &2955  &78765 \\
            &Tracktor~\cite{bergmann2019tracking}            & 56.2      & 54.9      & 44.6         &617       &\textbf{2394}  &76844 \\
            &ArTIST~\cite{saleh2021probabilistic}              & 56.6      & 57.8      & -         &519       &3532  &75031 \\
            &GSM~\cite{liu2020gsm} (G)               & 57.0      & 58.2      & 45.9        &475       &4332  &73573 \\
            &GNNMatch~\cite{papakis2020gcnnmatch} (G)            & 57.2      & 55.0      & 44.6         &559       &3905  &73493 \\
            &MPNTrack~\cite{braso2020learning} (G)$\star$   & \underline{58.6}      & \underline{61.7}      & \textbf{48.9}         &\textbf{354}       &4949  &\textbf{70252} \\
            &TADAM~\cite{guo2021online}               & \textbf{59.1}      & 59.5      & -         &529       &\underline{2540}  &71542 \\
            &Ours                & 58.3      & \textbf{62.1}      & \underline{48.6}      &\underline{389}        &4300  &\underline{71287} \\
\hline
MOT17     &DeepMOT~\cite{xu2020train}            & 53.7      & 53.8      & 42.4         &1947       &11731  &247447 \\
            &BLSM-MTP~\cite{kim2021discriminative}            & 55.9      & 60.4      & -         &\underline{1188}       &\textbf{8653}  &238853 \\
            &Tracktor~\cite{bergmann2019tracking}            & 56.3      & 55.1      & 44.8         &1987       &\underline{8866}  &235449 \\
            &GSM~\cite{liu2020gsm} (G)                & 56.4      & 57.8      & 45.7        &1485       &14379  &230174 \\
            &ArTIST~\cite{saleh2021probabilistic}              & 56.7      & 57.5      & -         &1756       &12353  &230437 \\
            &GNNMatch~\cite{papakis2020gcnnmatch} (G)           & 57.3      & 56.3      & 45.4         &1911       &14100  &225042 \\
            &MPNTrack~\cite{braso2020learning} (G)$\star$   & \underline{58.8}      & \textbf{61.7}      & \textbf{49.0}         &\textbf{1185}       &17413  &\textbf{213594} \\
            &TADAM~\cite{guo2021online}               & \textbf{59.7}      & 58.7      & -         &1930       &9676  &216029 \\
            &Ours                & 58.5      & \underline{61.2}      & \underline{48.3}      &1320       &17374  &\underline{215460} \\
\hline
MOT20       &Tracktor~\cite{bergmann2019tracking}            & 52.6      & 52.7      & 42.1             &1648       &\textbf{6930}  &236680 \\
            &ArTIST~\cite{saleh2021probabilistic}              & 53.6      & 51.0      & 41.6        &\underline{1531}       &\underline{7765}  &230576 \\
            &GNNMatch~\cite{papakis2020gcnnmatch} (G)            & 54.5      & 49.0      & 40.2         &2038       &9522  &223611 \\
            &TADAM~\cite{guo2021online}               & \underline{56.6}      & 51.6      & -         &1930       &39407  &\textbf{182520} \\
            &MPNTrack~\cite{braso2020learning} (G)$\star$   & \textbf{57.6}      & \textbf{59.1}      & \textbf{46.8}         &\textbf{1210}       &16953  &\underline{201384} \\
            &Ours                & 55.5     & \underline{55.4}      & \underline{43.6}      &1577       &10989  &217765 \\
\hline
\end{tabular}
}
\end{table}

\subsection{Comparison with Other Methods}
\label{sec:comparison}

Finally, we compare our method with other state-of-the-art approaches on the MOT16, MOT17 and MOT20 datasets. 
First, we test our method with the official public detections and compare the results with other public detection based methods. 
Second, because our method focuses on the association step, it can also be applied to any detector. Thus, we test our method on several other advanced private detectors based on MOT16 and MOT17 dataset to evaluate its generalization capacity. The results are shown in Table~\ref{tab:public_det_results} and Table~\ref{tab:private_det_results}.

\textbf{Public detections.} On the three tracking datasets, our method obtains consistent improvements compared with the baseline method Tracktor~\cite{bergmann2019tracking}. Specifically, we increase IDF1 by 7.2\%, 6.1\% and 2.7\%, respectively. ID switches also decrease by a large amount, indicating that our method performs better in association. Compared with other online methods, our approach also performs favorably. We achieve an IDF1 score 2.6\% to 8.6\% higher than the competitors on MOT16, 0.8\% to 12.5\% higher on MOT17 and 2.7\% to 6.4\% higher on MOT20. MOTA and IDS of our method are also better than those of most other methods. Our performance is even close to the offline method MPNTrack~\cite{braso2020learning}.
Our method causes more false positives (FP) than Tracktor. This is because forecasting the lost targets with a motion model inevitably causes some false detections. On the other hand, our forecasting helps to reduce additional false negative trajectories (FN) and in general improves MOTA by 2.1\%, 2.2\% and 2.9\%.


\begin{table}
\centering
\caption{Comparison with private detection based methods on the MOT16 (top) and MOT17 (bottom) test sets. G means graph based methods. $\uparrow$/$\downarrow$ denote that higher/lower results are better.}
\label{tab:private_det_results}
\resizebox{1.0\columnwidth}{!}{
\begin{tabular}{clllll}
\hline
Dataset &Method              & MOTA$\uparrow$      &IDF1$\uparrow$      & HOTA$\uparrow$      &IDS$\downarrow$       \\
\hline
\multirow{6}{*}{MOT16}     &DRT~\cite{wang2021drt}             & 71.1      & 60.9      & 50.7      &1173       \\

&Ours (Same detections as DRT)        & 71.2 (+0.1)     & 67.9 (+7.0)      & 54.2 (+3.5)      &1133 (-40)       \\
\cline{2-6}
&FUFET~\cite{shan2020tracklets} (G)              & 76.5      & 68.6      & 58.3      &1026       \\
&Ours (Same detections as FUFET)        & 76.7 (+0.2)     & 73.2 (+4.7)      & 59.7 (+1.4)      &705 (-321)       \\
\cline{2-6}
&CorrTracker~\cite{wang2021multiple} (G)              & 76.6      & 74.3      & 61.0      &979       \\
&Ours (Same detections as CorrTracker)        & 77.2(+0.6)      & 74.4 (+0.1)     & 60.6 (-0.4)      &740 (-239)      \\
\hline
\multirow{6}{*}{MOT17}&TrTrack~\cite{sun2020transtrack}             & 75.2      & 63.5      & 54.1         &3603       \\
&Ours (Same detections as TrTrack)      & 74.9 (-0.3)      & 67.2 (+3.7)      & 56.1 (+2.0)         &3948 (+345)       \\
\cline{2-6}
&GSDT~\cite{wang2021joint} (G)                & 73.2      & 66.5      & 55.2      &3891       \\
&Ours (Same detections as GSDT)         & 73.8 (+0.6)     & 71.8 (+5.3)     & 58.4 (+3.2)     &2586 (-1305)       \\
\cline{2-6}
&PermaTrackPr~\cite{tokmakov2021learning}                & 73.8      & 68.9      & 55.5      &3699       \\
&Ours (Same detections as PermaTrackPr)         & 74.5 (+0.7)     & 72.3 (+3.4)     & 57.3 (+1.8)      &2304 (-1395)      \\
\hline
\end{tabular}
}
\end{table}

\textbf{Private detections.} We also compare our approach with different private detection based methods to evaluate the generalization ability. Table~\ref{tab:private_det_results} shows the result of this comparison on the MOT16 and MOT17 test sets. We directly use the detection boxes of these trackers and associate with our Algorithm~\ref{alg:inference}. Our method, which only uses a simple graph neural network, obtains different degrees of improvements compared with these state-of-the-art methods. IDF1 of our method is 0.1\% to 7.0\% better than all of them, proving the strong association performance as well as the generalization capability of our method. For CorrTracker~\cite{wang2021multiple}, which constructed a extremely complex pyramid architecture and correlation learning, our method can still achieve slightly better results. 

Additionally, we compare our approach to other graph-based methods (labelled with "G") in Table~\ref{tab:public_det_results} and~\ref{tab:private_det_results}.
We consistently outperform the online competitors.
This demonstrates that properly preprocessing the available information to construct a suitable graph is crucial to achieve a robust tracking performance.

\section{Conclusion}

In this paper, we presented a new association method for multiple object tracking.
We analyze different feature integration methods, including a new IoU guided integration function, that are inserted into a simple message passing network to enhance the feature representations and overall  improve the robustness of existing trackers.
We show the importance and how to construct sparse graphs for the trajectory-detection association step, which reduces the computational cost and simultaneously improves the tracking performance notably.
Extensive experiments on publicly available datasets demonstrate that our method achieves state-of-the-art performance.
Additionally, our association method generalizes well and can also be applied to already strong performing tracking methods, resulting in further improved results.

\section*{Acknowledgement}
This work was supported in part by China Scholarship Council (NO. 201906840048). This work was also partially funded by the Austrian Research Promotion Agency (FFG) under the projects High-Scene (884306) and INTERACT (879648).

\bibliographystyle{named}
\bibliography{ijcai22}

\end{document}